\documentclass[conference]{IEEEtran}
\IEEEoverridecommandlockouts
\usepackage{cite}
\usepackage{amsmath,amssymb,amsfonts}
\usepackage{algorithmic}
\usepackage{graphicx}
\usepackage{textcomp}
\usepackage{xcolor}
\usepackage{float}
\usepackage{rotating}
\usepackage{ upgreek }
\usepackage{mathtools}
\usepackage{comment}
\usepackage{graphicx}
\usepackage{subfig}

\usepackage[normalem]{ulem}
\usepackage{diagbox}
\usepackage{multirow}
\usepackage{colortbl}

\begin{document}

\title{Towards an Ensemble Regressor Model for Anomalous ISP Traffic Prediction}

\author{\IEEEauthorblockN{Sajal Saha, Anwar Haque, and\ *Greg Sidebottom}
\IEEEauthorblockA{\textit{Department of Computer Science} \\
\textit{University of Western Ontario, London, ON, Canada}\\
\textit{*Juniper Networks, Kanata, ON, Canada}\\
Email:\{ssaha59, ahaque32\}@uwo.ca, *gsidebot@juniper.net}
}

\maketitle

\begin{abstract}
Prediction of network traffic behavior is significant for the effective management of modern telecommunication networks. However, the intuitive approach of predicting network traffic using administrative experience and market analysis data is inadequate for an efficient forecast framework. As a result, many different mathematical models have been studied to capture the general trend of the network traffic and predict accordingly. But the comprehensive performance analysis of varying regression models and their ensemble has not been studied before for analyzing real-world anomalous traffic. In this paper, several regression models such as Extra Gradient Boost (XGBoost), Light Gradient Boosting Machine (LightGBM), Stochastic Gradient Descent (SGD), Gradient Boosting Regressor (GBR), and CatBoost Regressor were analyzed to predict real traffic without and with outliers and show the significance of outlier detection in real-world traffic prediction. Also, we showed the outperformance of the ensemble regression model over the individual prediction model. We compared the performance of different regression models based on five different feature sets of lengths 6, 9, 12, 15, and 18. Our ensemble regression model achieved the minimum average gap of 5.04\% between actual and predicted traffic with nine outlier-adjusted inputs. In general, our experimental results indicate that the outliers in the data can significantly impact the quality of the prediction. Thus, outlier detection and mitigation assist the regression model in learning the general trend and making better predictions.
\end{abstract}


\section{Introduction}
Prediction of network traffic behavior in advance is immensely important for the effective management of modern telecommunication networks. It is equally vital for Internet Service Providers (ISP) for better resource management, efficient routing scheduling, short-term and long-term business capacity planning, advanced network design, and anomalous traffic detection and mitigation. An accurate traffic predictor framework can help ISP to manage their network proactively, which can ensure network Quality of Service (QoS) and Quality of Experience (QoE) \cite{2}. The modern network predictor tools predict and forecast traffic based on their historical data collected over a long time. This traffic information is captured using different network flow collection protocols, e.g., NetFlow \cite{rossi2010fine} and then converted the raw data into time-series format for further processing. The tools and techniques for processing temporal traffic information are extensively studied in the Time Series Forecasting domain \cite{wheelwright1998forecasting}. 


There are mainly two different types of mathematical models for analyzing network traffic, such as linear and non-linear model \cite{terui2002combined}. Linear models can learn patterns from those time series data where the next data point can be viewed as a linear combination of the past values. We can model the linear time series data using different statistical models AutoRegressive Integrated Moving Average (ARIMA), Seasonal ARIMA (SARIMA), Holt-Winter, etc. There are also linear machine learning models, e.g., linear regression, for analyzing the linear components of the time series data. But the real internet traffic is much more complex, having non-linear components challenging to capture using a linear model. In that case, non-linear models are deployed to forecast the real IP traffic. These models can be a statistical model such as Threshold AutoRegressive (TAR) \cite{ ricky2005threshold}, Exponential AutoRegressive (ExpAR) \cite{terui2002combined} etc. or machine learning model such as Lasso regression, Ridge regression, SVM (Support Vector Machine) with a non-linear kernel, Quantile regression, Bayesian Inference, etc. Also, we can use neural network-based machine learning models such as multi-resolution Finite-Impulse-Response (FIR) model \cite{ alarcon2006multiresolution}, Genetic Algorithm and Radial Based Function Network (GA-RBF) \cite{ wang2008internet } etc., deep learning model, e.g., DNN (Deep Neural Network), CNN(Convolutional Neural Network) or sequence model such RNN (Recurrent Neural Network), LSTM (Long Short Term Memory), etc.


Real-world IP network traffic is susceptible to various external and internal factors which may abruptly change the normal traffic flow. Internal factors are related to ISP companies, such as introducing new services, traffic migration, speed up-gradation, etc. In contrast, external factors are related to external ISP events and conditions such as new internet applications, regional economic factors, seasonal effects, etc. However, the unexpected suspicious changes, differing significantly from most data points, are prevalent in real IP traffic known as outliers or anomalies. The errors or problems in the traffic data collection sensor might be a source of those anomalous data points in the traffic. Those abnormal data points in traffic flow might impact learning the general trend in data. As a result, the prediction model might produce wrong inferences in the future, considering them as normal behavior. Therefore, it is essential to identify and handle the anomalies/outlier in the internet traffic before applying any prediction model. In addition, it might help the prediction model increase the generalization capability. Machine Learning models have been used to predict the complex real IP traffic, and they compared their performance with the statistical model or deep learning model. However, the comparative performance analysis among various regression models and stacking ensemble model to predict anomalous internet traffic has not been studied considerably. This work extensively analyzes the real-world IP traffic to identify the abnormal data points and mitigate them before feeding them into the regression model. We focus on the performance evaluation of the regression model and their ensemble models in IP traffic prediction with and without the outlier. Also, we experimented with a different feature set for training our machine learning model to find the optimal number of input features. Our experimental results show better accuracy when the machine learning models are trained using the anomaly-adjusted data. The main contributions of this work are:
 \begin{itemize}
    \item Outlier or anomaly detection using the standard statistical procedure and unsupervised learning method before using them to design our predictive model. 
    \item Finding the optimum number of features for the machine learning model. 
   \item Comparative performance analysis of different regression modeling techniques such as Extra Gradient Boosting, Light Gradient Boosting Machine (LightGBM), Cat Boosting (CatBoost), Stochastic Gradient Boosting (SGD), Gradient Boosting Regressor (GBR), and their stacking ensemble model with an outlier in the data. 
  \item Comparative performance analysis of the regression modeling techniques after adjusting the outliers in the data.  
\end{itemize}
This paper is organized as follows. Section \ref{literature review} describes the literature review of current traffic prediction using machine learning models. Section \ref{methodology} presents the methodology, including dataset description, machine learning models explanation, anomaly identification process, and experiment details. Section \ref{result} summarizes the different machine learning methods' performance and draws a comparative picture among prediction models with and without outliers in the dataset. Finally, section \ref{conclusion} concludes our paper and sheds light on future research directions.

\section{Literature Review}
\label{literature review}
S. Fischer et al. \cite{3} proposed a traffic prediction system called DEEPFLOW that processes the ingress traffic data and produces prediction for all traffic flows using different machine learning techniques. The prediction model includes two different categories of models such as statistical model and neural network based deep learning model. They specifically focused on handling traffic volatility using non-neural model called VAR (Vector Autoregression). The average model prediction error was around 10\% according to their experimental results which could have been improved. Their DEEPFLOW framework did not consider any regression machine learning model for the traffic prediction. 

D. Szosta et al. \cite{4} extensively investigated both machine learning classification and regression models for optical network traffic prediction. The proposed model considered four different supervised machine learning model tested on real traffic pattern collected from Seattle Internet Exchange Point. Total of five different datasets they used in their experiment to evaluate the prediction performance based on their proposed evaluation metric called Traffic Level Prediction Quality (TLPQ). The prediction models were trained on different feature set arranging from various window size and other feature such minute, day, and traffic values. Their experimental claims the outperformance of regression model over classification model in traffic prediction. Y. Xu et. al \cite{5} proposed a Gaussian Process (GP) based machine learning model for real world traffic prediction. They compared the model performance against the state-of-the-art seasonal ARIMA (SARIMA) and sinusoid superposition. The computational complexity of extracting optimal hypermeters for the prediction model has also been reduced from $O(n^3)$ to $O(n^2)$. The GP based machine learning model shows the average prediction of 3\% when predicting one-hour-ahead traffic and it’s increased to 5\% when the prediction length is extended to ten-hours-ahead.  

T. P. Oliveira et. al \cite{7} experimented with two different machine learning model such as multi-layer perceptron and stacked autoencoder for traffic prediction. They used a dataset collected from a private ISP in European cities.They used two hidden layers of MLP (Multi Layer Perceptron) with 60 and 40 neurons, respectively, while they found the best result for four hidden layers SAE with 80, 60, 60, and 40 neurons. They trained their model for 1000 epochs in both MLP and SAE (Stacked Auto Encoder), although the SAE training was divided into two-stage as the unsupervised pre-training for 900 epochs and 100 epochs supervised training. Different prediction length has been investigated using their prediction model and the error was increasing with longer prediction length. Their experimental results show the better performance of the simpler MLP than the SAE deep neural network.  Also, the MLP has taken lesser computational resource than SAE. D. Szostak et. al \cite{8} formulated traffic prediction problem into a classification problem and their proposed classification model predicts traffic bitrates level instead of exact traffic volume. They used Linear Discriminant Analysis (LDA) classifier to forecast the future traffic and it shows 93\% accuracy according to their experimental results. The experiment was conducted on the real data collected by Internet Exchange Point in Seattle. Y. Zang et. al \cite{10} author proposed a traffic prediction model by combining K-means clustering, Elman-NN, and wavelet decomposition. They cluster the adjacent and correlated base stations using K-means so it can improve the overall prediction accuracy. They reconstructed traffic into high-frequency and low-frequency components using wavelet decomposition for feeding to their traffic prediction model. Finally, a three-layer feed-forward neural network ENN is used to train the decomposed traffic data for making the prediction.

From the above discussion, it is apparent that the machine learning is extensively used for traffic prediction. Many works compared the performance of the machine learing model with neural network model and statistical model. However, we found a lack of investigation in the performance comparision among differnt regressor model and their ensemble model. Also, the real world internet traffic is very erratic in nature due to different external and internal factors which inject abnormal data points in the traffic flow. In this work, we identify and mitigate the outlier before using them to train our model and compare the results with data having outlier. To the best of our knowledge, this is the first work in the internet traffic prediction domain that offers a comparative view from different regressor and their ensemble model performance with anomaly detection, mitigation, and input optimization.

\section{Methodology}
\label{methodology}
First of all, we explained our data preprocessing module in subsection \ref{dataset} to handle the missing values and analyze traffic characteristics. Then, we described the data windowing phase for making the compatible for supervised learning in section \ref{Data Windowing and Feature Extraction}. The anomaly detection and mitigation module is explained in subsection \ref{anomaly_detection}. After that, we briefly described different regression model background in subsection \ref{model}. The model performance evaluation metrics are described in subsection \ref{metric}. Finally, we summarize the configuration of our experimental environment in subsection \ref{software}.

\subsection{Data Preprocessing}
\label{dataset}
Real internet traffic telemetry on several high-speed interfaces has been used for this experiment. The data are collected every five minutes for a recent thirty days time period. The data contains average generated traffic per five minutes in bit per second (bps). There are 8563 data samples in our dataset consisting of 29 days of complete data (288 data instances per day), while the last one-day data is incomplete. Only the timestamp (GMT) and traffic data (bit per second) are taken from the JSON file, and all other information is discarded. We removed this incomplete information from the original time series data for the experiment. Finally, the unit of our traffic data has been changed from bps (bit per second) to Gbps (gigabit per second) as the original value is large for feeding into our statistical model.

\subsection{Data Windowing and Feature Extraction}
\label{Data Windowing and Feature Extraction}
\begin{figure}[!htbp]
\centering
    \includegraphics[width=9cm,height = 2.5cm]{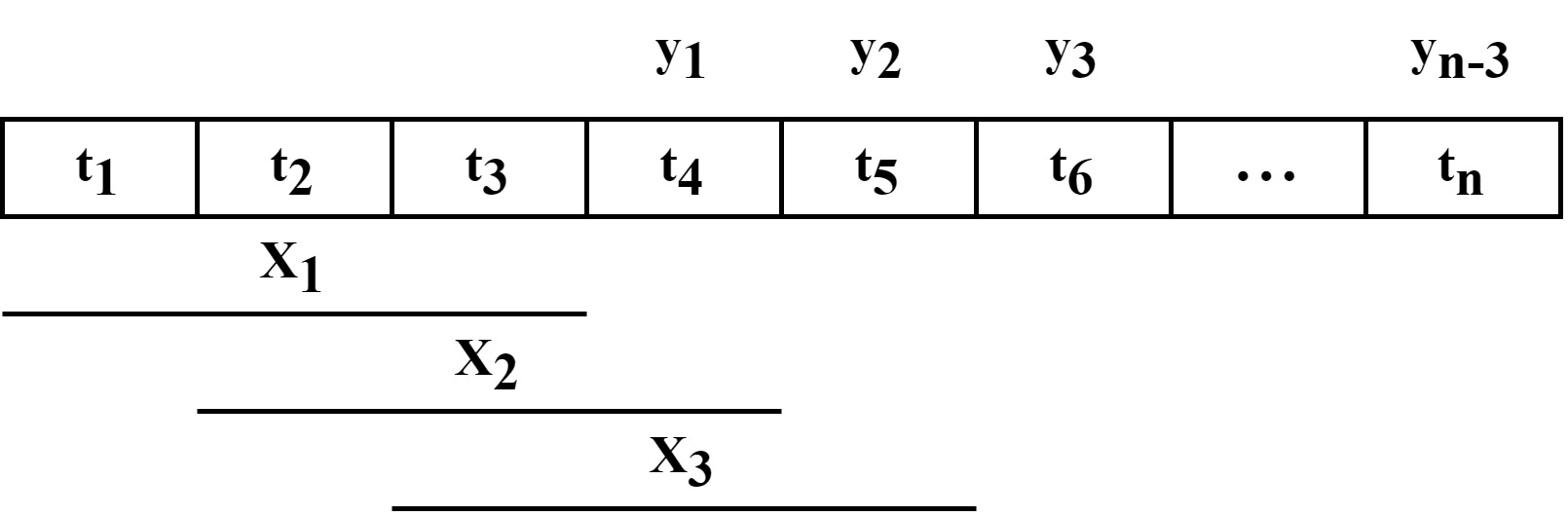}
    \caption{Data windowing}
    \label{fig:Data Windowing}
\end{figure}

Time series data need to be expressed into the proper format for supervised learning. Generally, the time-series data consists of several tuples (time, value), which is inappropriate for feeding them into the machine learning model. So, we restructured our original time series data using the sliding window technique. The sliding window technique is illustrated in Fig. \ref{fig:Data Windowing}. Three historical points ($X$) called features are used to predict the next data point ($y$) known as a target in the given sliding window example.

\begin{figure}
        \centering
        \includegraphics[width=9cm,height=4cm]{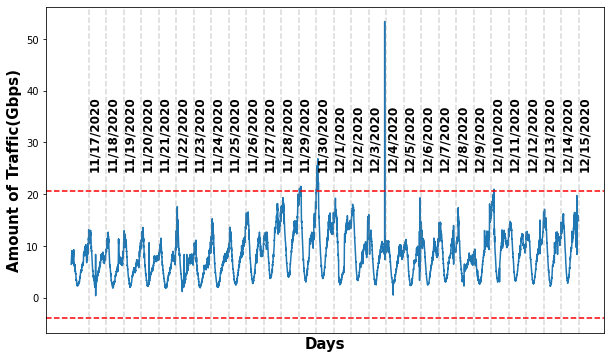}
        \caption{Outliers identified by Three-Sigma rule}
        \label{fig:three sigma}
\end{figure}
\begin{figure}
        \centering
        \includegraphics[width=9cm,height=4cm]{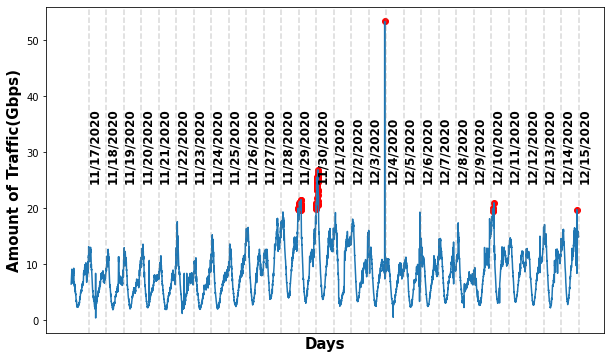}
        \caption{Outliers identified by Isolation Forest}
        \label{fig:isolation forest}
\end{figure}

\subsection{Anomaly detection and mitigation}
\label{anomaly_detection}
In time-series data, the anomaly or outlier are those data points significantly different from the general trend or pattern of the entire data series. The statistical properties of the outliers are not aligned with the other data points in the time series.  There are mainly three ways to identify outliers in the data set: predictive confidence level approach, statistical profiling approaches, and clustering-based unsupervised approach. Our experiment used statistical profiling and an unsupervised learning model to identify the outlier in our time series data. We used the standard Three-Sigma rule to extract the inlier data points within three standard deviations from a mean of the time series. All other data points exceeding the boundary are considered outliers or anomalies.  We also applied an unsupervised machine learning model called Isolation Forest to compare the performance of two different types of outlier detection techniques. This technique follows a recursive algorithm to isolate the data points by randomly selecting an attribute and their splitting value between the minimum and maximum value of the attribute.

\begin{table*}[!ht]
\caption{Traffic prediction model performance summary}
\label{tab:prediction_summary}
\centering
\begin{tabular}{|l|l|l|l|l|l||l|l|l|l|l|} 
\hline
\diagbox{}{Result}        & \multicolumn{5}{c||}{MAPE(\%)[Data with outlier]}                                                                             & \multicolumn{5}{c|}{MAPE(\%) [Data without outlier]}                                                                           \\ 
\cline{2-11}
\diagbox{Model}{Input} & \multicolumn{1}{c|}{6} & \multicolumn{1}{c|}{9} & \multicolumn{1}{c|}{12} & \multicolumn{1}{c|}{15} & \multicolumn{1}{c||}{18} & \multicolumn{1}{c|}{6} & \multicolumn{1}{c|}{9} & \multicolumn{1}{c|}{12} & \multicolumn{1}{c|}{15} & \multicolumn{1}{c|}{18}  \\ 
\hline
XGBoost                       & 7.65                   & 7.69                   & \textbf{\uline{7.47}}   & 7.59                    & 7.60                     & 5.16                   & 5.20                   & 5.19                    & \textbf{\uline{5.15}}   & 5.17                     \\ 
\hline
LightGBM                      & 8.51                   & 8.47                   & \textbf{\uline{8.47}}   & 8.47                    & 8.53                     & 5.11                   & \uline{\textbf{5.09}}  & 5.13                    & 5.14                    & 5.10                     \\ 
\hline
SGD                           & 12.80                  & 10.51                  & \textbf{\uline{10.44}}  & 11.13                   & 12.16                    & \textbf{\uline{6.10}}  & 7.27                   & 8.01                    & 8.50                    & 8.23                     \\ 
\hline
GBR                           & \textbf{\uline{7.47}}  & 7.74                   & 7.57                    & 7.60                    & 7.76                     & \textbf{\uline{5.20}}  & 5.31                   & 5.24                    & 5.26                    & 5.25                     \\ 
\hline
CatBoost                      & \textbf{\uline{7.56}}  & 7.58                   & 7.64                    & 7.78                    & 8.12                     & 5.10                   & \textbf{\uline{5.08}}  & 5.20                    & 5.32                    & 5.44                     \\ 
\hline
Ensemble                      & 7.36                   & 7.47                   & \textbf{\uline{7.23}}   & 8.07                    & 7.86                     & 5.05                   & \textbf{\uline{5.04}}  & 5.08                    & 5.07                    & 5.10                     \\
\hline
\end{tabular}
\end{table*}

\subsection{Prediction models}
\label{model}
\subsubsection{Extreme Gradient Boosting (XGBoost)}
It is an implementation of of the gradient boosting algorithm initially developed in \cite{2016}. This model can be used for both classification and regression problem, and it is comparatively fast in computation in comparison with the other implementation of gradient boosting. 


\subsubsection{Light Gradient Boosting Machine (LightGBM)} LightGBM is another implementation of gradient boosting algorithm proposed in \cite{ke2017lightgbm}. This framework minimize the limitation of the histogram-based algorithm by introducing two novel techniques: Gradient-based One Side Sampling and Exclusive Feature Bundling (EFB).


\subsubsection{Cat Boost Regressor (CatBoost)} This is a binary-tree based implementation of gradient boosting. CatBoost technique address a very general implementation problem of gradient boosting and tries to solve the issue by proposing ordering principle. All of the gradient boosting implementation relies on the targets of all training samples after several steps \cite{prokhorenkova2017catboost}. 


\subsubsection{Stochastic Gradient Descent (SGD)} It is a simple but efficient machine learning model used for both classification and regression \cite{saad1998online}. The basic difference between gradient descent and SGD is the number of samples taken to compute the derivatives. SGD randomly select one data sample in each iteration for calculating the gradient which significantly reduces the number of computation in comparison with the gradient descent.

\subsubsection{Gradient Boosting Regressor (GBR)} 
Gradient Boosting permits for the optimization of random differentiable loss functions and constructs an additive model in a forward stage-wise process \cite{scikit-learn}.The non-positive gradient of the provided loss function is fitted by a regression tree in each stage.

\subsubsection{Stacked Generalization Model}
Stacked Generalization is an ensemble technique of combining the results from different individual models for improving the performance. The individual predictors are called Level-$0$ or base models while the meta-model that integrates the outcomes of the base models, referred to as a level-$1$ model.

\subsection{Evaluation Metrics}
\label{metric}
We used Mean Absolute Percentage Error (MAPE) to estimate the performance of our traffic forecasting models. The performance metric identifies the deviation of the predicted result from the original data. For example, MAPE error represents the average percentage of fluctuation between the actual value and predicted value. Therefore, we can define our performance metric mathematically as follow: 
\begin{equation}
    MAPE = \dfrac{1}{n}\sum_{i=1}^n \bigg| \dfrac{p_i-o_i}{o_i} \bigg| \times 100 \%
\end{equation}
\begin{itemize}
    \item Here, $p_i$ and $o_i$ are predicted and original value respectively
    \item $n$ is the total number of test instance
\end{itemize}
\subsection{Software and Hardware Preliminaries}
\label{software}
We used Python and machine learning library Scikit-Learn\cite{scikit-learn} to conduct the experiments.  Our computer has the configuration of Intel (R) i3-8130U CPU@2.20GHz, 8GB memory, and a 64-bit Windows operating system.




\section{Results and Discussion}
\label{result}
Our experiment considered 21 days of IP traffic to train our machine learning models, while the last eight days of traffic were used for testing. Our experiment has mainly two phases: I) anomaly detection and mitigation, and II) machine learning model implementation and performance evaluation. 


Two different methods were applied to determine the abnormal data points in our traffic in phase I. The Three-Sigma rule identifies those data points as outliers which are three standards that deviated from the mean of the traffic data. Fig. \ref{fig:three sigma} illustrates the inlier points wrapped by the upper and lower horizontal red line. The data points lie outside the top, and bottom red line boundary in \ref{fig:three sigma} are identified as outliers according to the Three-Sigma rule.

\begin{figure*} [!htbp]
\centering
\subfloat[Predicted traffic by Ensemble model with outlier in the data]{\includegraphics[width=9cm,height=4cm]{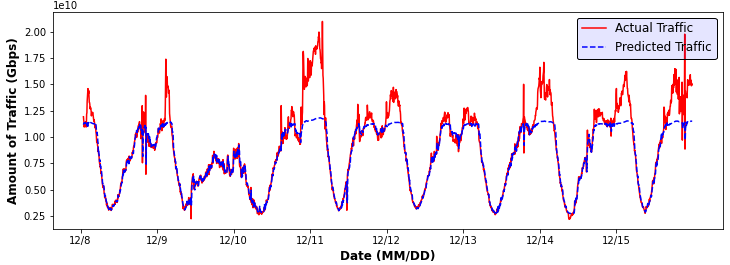}}
\subfloat[Predicted traffic by Ensemble model without outlier in the data]{\includegraphics[width=9cm,height=4cm]{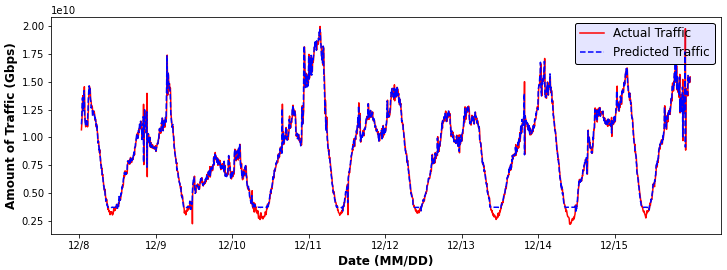}}
\caption{Actual traffic vs predicted traffic by best performing model}
\label{fig:Comparative result analysis}
\end{figure*}

\begin{figure*} [!htbp]
\centering
\subfloat[Accuracy comparison with outlier in the data]{\includegraphics[width=2.5in,height=3.5in]{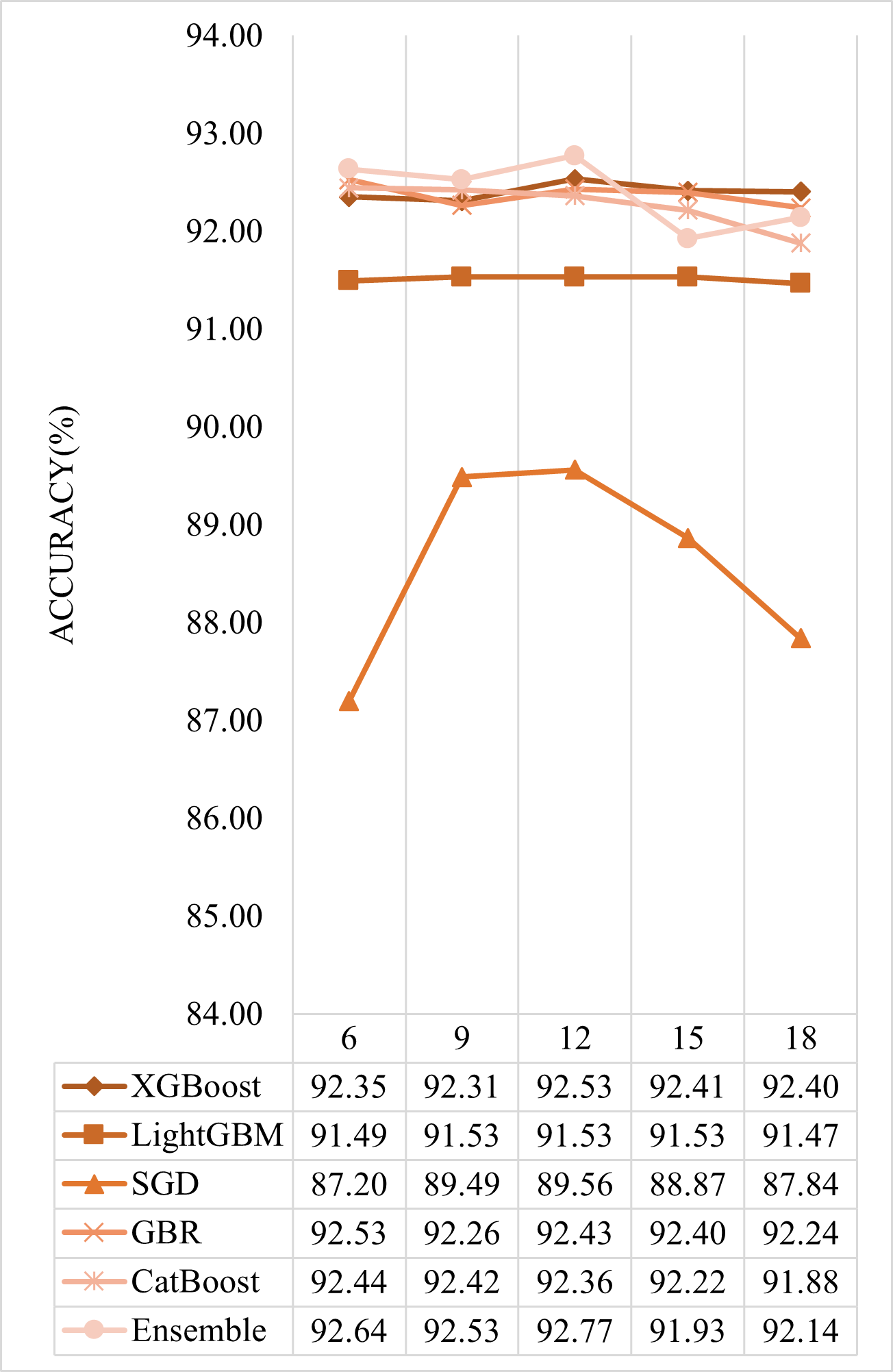}}
\hspace{1cm}
\subfloat[Accuracy comparison without outlier in the data]{\includegraphics[width=2.5in,height=3.5in]{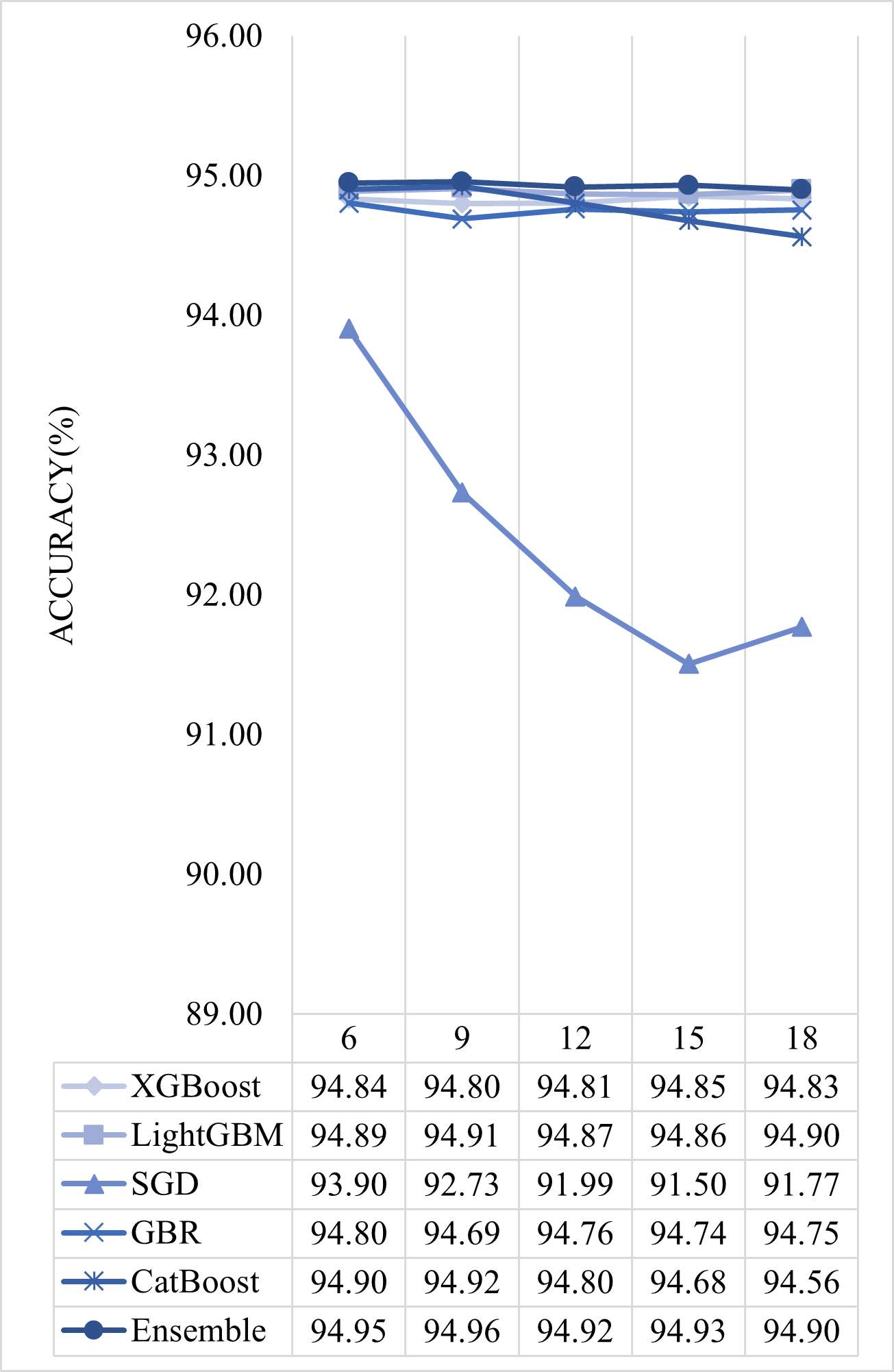}}
\caption{Model accuracy comparison for different input length}
\label{fig:accuracy comparision}
\end{figure*}

Furthermore, an unsupervised learning algorithm called Isolation Forest is also used to identify the outliers, and the result is presented in Fig. \ref{fig:isolation forest}. From \ref{fig:three sigma} and \ref{fig:isolation forest}, a similar patterns of outliers are identified in our traffic data. We adjusted the outliers identified by the Three-Sigma rule before using them to train our prediction models. The technique used for adjusting the outliers is called backward filling, in which the following valid data point replaces the outlier.

Finally, in phase II, we applied several machine learning models such as XGBoost, LightGBM, CatBoost, SGD, GBR, and an ensemble model based on stacking technique. We used five-fold cross-validation to train our models. The performance evaluation metrics of our experimental model are summarized in Table \ref{tab:prediction_summary}. Two different versions of each model were investigated based on the data with and without outlier. We used a total of five different feature sets of equivalent traffic length of 30 minutes, 45 minutes, 60 minutes, 75 minutes, and 90 minutes for training our model to identify the optimum number of inputs for our prediction model. The best results for the individual model are marked bold and underlined in the table.

For the XGBoost model, we achieved the best prediction using 12 and 15 inputs respectively for model with and without outlier data. The best average deviation between actual and predicted traffic is 7.47\% and 5.15\% with and without outlier, respectively, which indicated more than 30\% improvement in traffic prediction after adjusting outliers.  In the LightGBM model, the minimum prediction error between actual and predicted traffic is 5.09\% for 9 outlier-adjusted input variables.The prediction improvement is around 40\% when compared with the best forecasting error of 8.51\% for the prediction model having outliers in the training data. For other input variables 9, 12, 15, and 18, we have seen a similar amount, around 40\% (8.51\% to 5.11\%, 8.47\% to 5.13\%, 8.47\% to 5.14\%, and 8.53\% to 5.10\% respectively) of prediction enhancements after adjusting the anomalies in the data. For SGD, we found the best prediction result with MAPE of 6.10\% using input length of 6, and the result improved by more than 41\% after removing the outliers compared with 10.44\% error in prediction with outlier data. . The other input variables show a lower gap between actual and predicted traffic for outlier-adjusted data. The GBR model showed the minimum deviation of 5.20\% when trained using thirty minutes of traffic data without the outlier. After mitigating abnormal data points, our experimental results depict more than 30\% of better prediction for all input variables. In CatBoost model, the minimum prediction error of 5.08\% is achieved using 9 input variables. The traffic prediction improvement for the CatBoost model is more than 30\% compared with the minimum error of 7.56\% in the case of outlier data. Finally, we ensemble the individual prediction model based on the stacking technique to further reduce the gap between actual and predicted traffic. According to our experimental result, the ensemble model is the best prediction model with a minimum prediction error of 7.23\% (12 input variables) with outliers in the data and 5.04\% (9 input variables) without outliers. Our ensemble model provided the lowest error amongst all the prediction models in both training strategies. A graphical representation of actual traffic and predicted traffic using the best performing ensemble model with and without outliers are shown in Fig. \ref{fig:Comparative result analysis}(a) and Fig. \ref{fig:Comparative result analysis}(b). Our experimental results show overall performance improvement for all considering machine learning models after adjusting the abnormal traffic. From our experimental result we can conclude, the ourlier mitigation in the dataset before trining the prediction model can improve the prediction by average 30\% more accuracy. Our experimental result shows that outlier mitigation in the dataset before training the prediction model can improve the prediction by an average of 30\% more accuracy. 

The comparative analysis of the model accuracy (100-MAPE) based on different input lengths are depicted in Fig. \ref{fig:accuracy comparision}. Our experimental results showed that the performance of SGD is more sensitive to the input lengths. The SGD model accuracy rises from input 6 to 9 then plummets from 15 to 18 when trained with outlier data. But in the other case, the accuracy decreases with the increased input length. There is a variation in the model accuracy with the input length for other models, but the magnitude is much lower than the SGD. 

However, many internal and external factors can affect the regular traffic pattern in the real world. Since machine learning-based traffic prediction algorithms learn the general pattern in the dataset and predict accordingly, it is essential to handle the outliers before providing them to learn. Otherwise, there is a chance of learning from abnormal traffic patterns, affecting the prediction result. Our experimental results also showed that outliers in the data make the model performance poor than the clean data. Our results also showed the out-performance of the ensemble machine learning model than the individual model. Moreover, we noticed a varying model performance for different input lengths. Therefore, it is also crucial to optimize input size for the prediction model.  

\section{Conclusion}
\label{conclusion}
In this work, we experimented with several regression models in internet traffic prediction. A total of five individual  regressor model such as XGBoost, LightGBM, SGD, GBR, CatBoost and one ensemble model were implemented to analyze our real-world traffic. We applied these models with and without outliers in traffic to demonstrate the impact of anomalies in traffic prediction. Also we demonstrated the outperformance of ensemble model than individual regressor model. Our experimental results show a significant performance improvement for all models after adjusting the abnormal data points in the original dataset. The ensemble model shows the minimum deviation between actual and predicted traffic with and without outlier in the data. In the real world, the internet traffic is very susceptible to the different external and internal factors which make the traffic non-linear and difficult to predict. The sudden changes in the data should be treated first before providing them into a machine learning model to predict. Otherwise, the model will be unable to generalize the actual traffic pattern, and predictions will not be effective. In the future, we would like to extend this work by multi-step traffic prediction. Also, we would explore different unsupervised methods in real-time anomaly detection and mitigation.
\bibliographystyle{IEEEtran}
\bibliography{conference_101719}

\begin{thebibliography}{10}
\providecommand{\url}[1]{#1}
\csname url@samestyle\endcsname
\providecommand{\newblock}{\relax}
\providecommand{\bibinfo}[2]{#2}
\providecommand{\BIBentrySTDinterwordspacing}{\spaceskip=0pt\relax}
\providecommand{\BIBentryALTinterwordstretchfactor}{4}
\providecommand{\BIBentryALTinterwordspacing}{\spaceskip=\fontdimen2\font plus
\BIBentryALTinterwordstretchfactor\fontdimen3\font minus
  \fontdimen4\font\relax}
\providecommand{\BIBforeignlanguage}[2]{{%
\expandafter\ifx\csname l@#1\endcsname\relax
\typeout{** WARNING: IEEEtran.bst: No hyphenation pattern has been}%
\typeout{** loaded for the language `#1'. Using the pattern for}%
\typeout{** the default language instead.}%
\else
\language=\csname l@#1\endcsname
\fi
#2}}
\providecommand{\BIBdecl}{\relax}
\BIBdecl

\bibitem{2}
S.~Troia, R.~Alvizu, Y.~Zhou, G.~Maier, and A.~Pattavina, ``Deep learning-based
  traffic prediction for network optimization,'' in \emph{2018 20th
  International Conference on Transparent Optical Networks (ICTON)}.\hskip 1em
  plus 0.5em minus 0.4em\relax IEEE, 2018, pp. 1--4.

\bibitem{rossi2010fine}
D.~Rossi and S.~Valenti, ``Fine-grained traffic classification with netflow
  data,'' in \emph{Proceedings of the 6th international wireless communications
  and mobile computing conference}, 2010, pp. 479--483.

\bibitem{wheelwright1998forecasting}
S.~Wheelwright, S.~Makridakis, and R.~J. Hyndman, \emph{Forecasting: methods
  and applications}.\hskip 1em plus 0.5em minus 0.4em\relax John Wiley \& Sons,
  1998.

\bibitem{terui2002combined}
N.~Terui and H.~K. Van~Dijk, ``Combined forecasts from linear and nonlinear
  time series models,'' \emph{International Journal of Forecasting}, vol.~18,
  no.~3, pp. 421--438, 2002.

\bibitem{ricky2005threshold}
B.~Ricky~Rambharat, A.~E. Brockwell, and D.~J. Seppi, ``A threshold
  autoregressive model for wholesale electricity prices,'' \emph{Journal of the
  Royal Statistical Society: Series C (Applied Statistics)}, vol.~54, no.~2,
  pp. 287--299, 2005.

\bibitem{alarcon2006multiresolution}
V.~Alarcon-Aquino and J.~A. Barria, ``Multiresolution fir neural-network-based
  learning algorithm applied to network traffic prediction,'' \emph{IEEE
  Transactions on Systems, Man, and Cybernetics, Part C (Applications and
  Reviews)}, vol.~36, no.~2, pp. 208--220, 2006.

\bibitem{wang2008internet}
C.~Wang, X.~Zhang, H.~Yan, and L.~Zheng, ``An internet traffic forecasting
  model adopting radical based on function neural network optimized by genetic
  algorithm,'' in \emph{First International Workshop on Knowledge Discovery and
  Data Mining (WKDD 2008)}.\hskip 1em plus 0.5em minus 0.4em\relax IEEE, 2008,
  pp. 367--370.

\bibitem{3}
S.~Fischer, K.~Katsarou, and O.~Holschke, ``Deepflow: Towards network-wide
  ingress traffic prediction using machine learning at large scale,'' in
  \emph{2020 International Symposium on Networks, Computers and Communications
  (ISNCC)}.\hskip 1em plus 0.5em minus 0.4em\relax IEEE, 2020, pp. 1--8.

\bibitem{4}
D.~Szostak, A.~W{\l}odarczyk, and K.~Walkowiak, ``Machine learning
  classification and regression approaches for optical network traffic
  prediction,'' \emph{Electronics}, vol.~10, no.~13, p. 1578, 2021.

\bibitem{5}
Y.~Xu, W.~Xu, F.~Yin, J.~Lin, and S.~Cui, ``High-accuracy wireless traffic
  prediction: A gp-based machine learning approach,'' in \emph{GLOBECOM
  2017-2017 IEEE Global Communications Conference}.\hskip 1em plus 0.5em minus
  0.4em\relax IEEE, 2017, pp. 1--6.

\bibitem{7}
T.~P. Oliveira, J.~S. Barbar, and A.~S. Soares, ``Multilayer perceptron and
  stacked autoencoder for internet traffic prediction,'' in \emph{IFIP
  International conference on network and parallel computing}.\hskip 1em plus
  0.5em minus 0.4em\relax Springer, 2014, pp. 61--71.

\bibitem{8}
D.~Szostak, K.~Walkowiak, and A.~W{\l}odarczyk, ``Short-term traffic
  forecasting in optical network using linear discriminant analysis machine
  learning classifier,'' in \emph{2020 22nd International Conference on
  Transparent Optical Networks (ICTON)}.\hskip 1em plus 0.5em minus 0.4em\relax
  IEEE, 2020, pp. 1--4.

\bibitem{10}
Y.~Zang, F.~Ni, Z.~Feng, S.~Cui, and Z.~Ding, ``Wavelet transform processing
  for cellular traffic prediction in machine learning networks,'' in \emph{2015
  IEEE China Summit and International Conference on Signal and Information
  Processing (ChinaSIP)}.\hskip 1em plus 0.5em minus 0.4em\relax IEEE, 2015,
  pp. 458--462.

\bibitem{2016}
\BIBentryALTinterwordspacing
T.~Chen and C.~Guestrin, ``Xgboost,'' \emph{Proceedings of the 22nd ACM SIGKDD
  International Conference on Knowledge Discovery and Data Mining}, Aug 2016.
  [Online]. Available: \url{http://dx.doi.org/10.1145/2939672.2939785}
\BIBentrySTDinterwordspacing

\bibitem{ke2017lightgbm}
G.~Ke, Q.~Meng, T.~Finley, T.~Wang, W.~Chen, W.~Ma, Q.~Ye, and T.-Y. Liu,
  ``Lightgbm: A highly efficient gradient boosting decision tree,''
  \emph{Advances in neural information processing systems}, vol.~30, pp.
  3146--3154, 2017.

\bibitem{prokhorenkova2017catboost}
L.~Prokhorenkova, G.~Gusev, A.~Vorobev, A.~V. Dorogush, and A.~Gulin,
  ``Catboost: unbiased boosting with categorical features,'' \emph{arXiv
  preprint arXiv:1706.09516}, 2017.

\bibitem{saad1998online}
D.~Saad, ``Online algorithms and stochastic approximations,'' \emph{Online
  Learning}, vol.~5, pp. 6--3, 1998.

\bibitem{scikit-learn}
F.~Pedregosa, G.~Varoquaux, A.~Gramfort, V.~Michel, B.~Thirion, O.~Grisel,
  M.~Blondel, P.~Prettenhofer, R.~Weiss, V.~Dubourg, J.~Vanderplas, A.~Passos,
  D.~Cournapeau, M.~Brucher, M.~Perrot, and E.~Duchesnay, ``Scikit-learn:
  Machine learning in {P}ython,'' \emph{Journal of Machine Learning Research},
  vol.~12, pp. 2825--2830, 2011.

\end{thebibliography}

\vspace{12pt}

\end{document}